\documentclass[10pt,twocolumn,letterpaper]{article}

\usepackage{cvpr}      

\definecolor{cvprblue}{rgb}{0.21,0.49,0.74}
\usepackage[pagebackref,breaklinks,colorlinks,allcolors=cvprblue]{hyperref}
\usepackage{multirow}

\def\paperID{14290} 
\def\confName{CVPR}
\def\confYear{2025}

\title{KAN not Work: Investigating the Applicability of Kolmogorov-Arnold Networks in Computer Vision
}

\author{
    Yueyang Cang\textsuperscript{1}\thanks{Equal contribution.} \quad Yuhang Liu\textsuperscript{1}\footnotemark[1] \quad Shi Li\textsuperscript{1}\thanks{Corresponding author: Li Shi (\texttt{shilits@tsinghua.edu.cn})} \\[1ex]
    \textsuperscript{1}Tsinghua University, Beijing, China \\
    {\tt\small cangyy23@mails.tsinghua.edu.cn, yh-liu23@mails.tsinghua.edu.cn, shilits@tsinghua.edu.cn}
}

\begin{document}
\maketitle
\begin{abstract}
Kolmogorov-Arnold Networks (KAN) have gained attention for their potential in capturing complex patterns, but the applicability of KAN and its variant, Convolutional KAN (CKAN), in computer vision remains a subject of significant debate. This study analyzes the performance of KAN and CKAN in computer vision tasks, with a focus on their robustness across different data scales and noise levels. The results show that CKAN, due to its extremely low generalization ability, often performs worse than traditional CNNs. Although the original KAN demonstrates strong fitting capabilities, its generalization ability is also weak, making it unsuitable for replacing the final layer of a model with an NLP layer. To address this issue, we propose a smoothness regularization method and a "Segment Deactivation" technique, both of which significantly improve the generalization ability of KAN, leading to enhanced performance when replacing the final NLP layer.
\end{abstract}
    
\section{Introduction}
\label{sec:intro}
In recent years, neural network\cite{hopfield1982neural,haykin2009neural,lecun2015deep} architectures have achieved remarkable breakthroughs in computer vision, especially in tasks such as image classification, and semantic segmentation\cite{krizhevsky2012imagenet,long2015fully,redmon2016you}. Traditional models, particularly Convolutional Neural Networks (CNNs)\cite{lecun1989backpropagation,krizhevsky2012imagenet,simonyan2014very,he2016deep,huang2017densely} and the more recent Transformer architectures\cite{vaswani2017attention,carion2020end,dosovitskiy2020image,carion2020end,liu2021swin,wang2021pyramid,yuan2021tokens,cheng2021maskformer}, have been widely adopted in these areas due to their ability to capture spatial hierarchies and long-range dependencies. CNNs leverage translational invariance through convolutional filters, while Transformers effectively handle complex patterns in images using self-attention mechanisms. These models have thus become mainstream approaches in computer vision tasks.

Alongside these established models, Kolmogorov-Arnold Networks (KAN)\cite{liu2024kan} have emerged as an intriguing alternative due to their theoretical efficiency and compact design. KAN are inspired by the Kolmogorov-Arnold representation theorem\cite{kolmogorov1956representation,kolmogorov1957superposition}, which suggests that any continuous multivariate function can be represented by a sum of univariate functions. Based on this principle, KAN utilize learnable activation functions, such as B-spline functions\cite{sprecher2002space,schoenberg1988contributions}, allowing them to adapt flexibly to complex patterns with fewer parameters compared to traditional networks. This characteristic makes KAN theoretically appealing for applications where model interpretability and parameter efficiency are prioritized, thus attracting significant research interest\cite{vaca2024kolmogorov,genet2024temporal,bozorgasl2024wav,bodner2024convolutional,bresson2024kagnns}.

This study aims to systematically evaluate the performance of KAN and its variants across several core computer vision tasks to explore their potential and limitations in vision applications, including image classification and semantic segmentation. In our experiments, we assessed the performance of KAN and Convolutional KAN\cite{bodner2024convolutional} models under varying data scales and noise levels. The results show that although KAN and Convolutional KAN exhibit strong fitting capabilities theoretically, their high sensitivity to noise severely impacts their robustness in vision tasks, leading to poor performance in practical vision applications.

To enhance KAN stability in vision tasks, we propose a novel smoothness regularization method designed to mitigate excessive oscillations in model parameters. Specifically, this regularization constrains the rate of change in spline functions, ensuring that the KAN model maintains smooth transitions when learning complex patterns, thus enhancing model stability and preventing overfitting. Additionally, we introduce a new technique called Segment Deactivation, which, during training, probabilistically reduces specific spline segments to linear functions. This approach effectively enhances the robustness of the training process by simplifying the model’s complexity, leading to improved performance of KANs on high-dimensional vision tasks.

To summarize, our Contributions Are as Follows:

\textbf{1. Evaluation of KANs in Vision Tasks.}  
We evaluated the performance of Kolmogorov-Arnold Networks (KAN) and its variants in key computer vision tasks, including image classification and segmentation, and identified significant generalization issues in their application to computer vision tasks.

\textbf{2. Smoothness Regularization.}  
To address KANs' sensitivity to noise, we propose a smoothness regularization method that stabilizes the model, enhancing robustness in noisy environments.

\textbf{3. Segment Deactivation.}  
We introduce Segment Deactivation, a technique that reduces certain spline segments to linear functions, improving KAN stability and adaptability to vision tasks.


\section{Related Work}
\label{sec:formatting}


\subsection*{Kolmogorov-Arnold representation theorem}

The Kolmogorov-Arnold representation theorem, proposed by mathematicians Andrey Kolmogorov and Vladimir Arnold, states that any continuous multivariate function \( f(x) \) can be represented as a sum of specific types of series, known as Kolmogorov series, composed of univariate continuous functions. For a continuous function \( f: [0,1]^d \rightarrow \mathbb{R} \), the theorem allows it to be decomposed into a finite combination of univariate functions as follows:
\begin{equation}
    f(x_1, \dots, x_d) = \sum_{q=1}^{2d+1} \Phi_q \left( \sum_{p=1}^d \varphi_{q,p}(x_p) \right)
\end{equation}

where \( \Phi_q \) and \( \varphi_{q,p} \) are continuous univariate functions. The importance of the Kolmogorov-Arnold representation theorem lies in its demonstration that any multivariate continuous function can be represented by a finite combination of univariate functions, significantly simplifying the representation and computation of complex functions. This property provides theoretical support for high-dimensional data modeling and has inspired neural network architectures, especially in the areas of function approximation and dimensionality reduction.

\subsection*{Kolmogorov-Arnold Networks}

Kolmogorov-Arnold Networks (KAN) are a neural network architecture inspired by the Kolmogorov-Arnold representation theorem. KAN approximates multivariate functions by decomposing them into combinations of univariate transformations and linear mappings. Specifically, KAN employs nonlinear spline-based activation functions, such as B-splines, to approximate complex functions with high-dimensional data inputs effectively. KAN has demonstrated promising results in function approximation and high-dimensional data modeling, leveraging its capacity for powerful, flexible function representation.

\subsection*{Convolutional KAN}

Convolutional KAN (CKAN) was introduced shortly after the development of Kolmogorov-Arnold Networks (KAN) to enhance their performance in visual tasks. By incorporating convolutional layers, CKAN combines the spatial feature extraction capabilities of Convolutional Neural Networks (CNNs) with the nonlinear approximation power of KAN's spline-based activation functions. In CKAN, the integration of KAN within convolutional layers enables the model to effectively capture spatial hierarchies in images while leveraging the expressive power of spline functions for complex nonlinear transformations. This approach allows CKAN to perform well in various vision tasks by unifying the advantages of CNNs in local feature extraction with KAN’s advanced representational flexibility.

\section{Method}

\subsection{Kolmogorov-Arnold Networks (KAN)}

Kolmogorov-Arnold Networks (KAN) are neural architectures derived from the Kolmogorov-Arnold representation theorem, which establishes that any continuous multivariate function can be represented by a sum of univariate functions. Leveraging this theorem, KAN decomposes high-dimensional inputs into univariate transformations, allowing effective approximation of complex multivariate functions.

Given an input vector \(\mathbf{x} = (x_1, x_2, \dots, x_{d_{\text{in}}})\) with dimension \(d_{\text{in}}\), KAN models the function \(f: \mathbb{R}^{d_{\text{in}}} \rightarrow \mathbb{R}^{d_{\text{out}}}\) as follows:

\begin{equation}
f(\mathbf{x}) = \Phi \circ \mathbf{x} = \left[ \sum_{i=1}^{d_{\text{in}}} \varphi_{i,1}(x_i), \dots, \sum_{i=1}^{d_{\text{in}}} \varphi_{i,d_{\text{out}}}(x_i) \right]
\end{equation}

where each \(\varphi_{i,j}\) represents a univariate transformation on input component \(x_i\), and \(\Phi\) encapsulates the composite mapping applied across input dimensions. KAN typically utilizes \textbf{B-spline functions} as activation functions due to their flexibility and smoothness, providing an enhanced capacity to capture nonlinearity.

The process of function approximation in KAN consists of several key steps:

\begin{enumerate}
    \item \textbf{Univariate Transformations}: Each input \(x_p\) undergoes transformation through a univariate function \(\varphi_{q,p}(x_p)\), producing intermediate values \(h_q\):
    \begin{equation}
    h_q = \sum_{p=1}^{d} \varphi_{q,p}(x_p)
    \end{equation}

    \item \textbf{Nonlinear Activation with B-splines}: Each summation result \(h_q\) is then passed through a nonlinear activation, formulated as a combination of the \textbf{SiLU (Sigmoid Linear Unit)} and a B-spline function:
    \begin{equation}
    h_q' = \phi_q(h_q) = w_{\text{b}} \cdot \text{silu}(h_q) + w_{\text{s}} \cdot \text{spline}(h_q)
    \end{equation}
    where \(\text{silu}(x) = \frac{x}{1 + e^{-x}}\) and \(\text{spline}(x) = \sum_i c_i B_i(x)\), with \(B_i(x)\) as the basis functions of the B-spline.

    \item \textbf{Final Output Computation}: The network aggregates the transformed hidden layer outputs to yield the final function approximation:
    \begin{equation}
    y = \sum_{q=1}^{2d+1} \phi_q \left( \sum_{p=1}^{d} \varphi_{q,p}(x_p) \right)
    \end{equation}
\end{enumerate}

This architecture effectively represents high-dimensional functions by leveraging the spline-based transformations, with SiLU and B-spline activations ensuring smooth, nonlinear responses. The univariate mapping strategy aligns with the Kolmogorov-Arnold theorem, enabling KAN to achieve accurate approximations for complex function classes.

\subsection{Convolutional KAN (CKAN)}

Convolutional Kolmogorov-Arnold Networks (CKAN) extend KAN by embedding its spline-based transformations into convolutional architectures, thus enhancing the network’s suitability for visual tasks involving spatially structured data. CKAN fuses convolutional operations with KAN’s expressive activation, optimizing it for capturing local image structures and complex non-linear interactions.

The specific architecture of CKAN is as follows:

\begin{enumerate}
    \item \textbf{Convolutional Transformation}: The network begins with convolutional layers, which perform spatial filtering on the input data. For each input feature map \(\mathbf{x}_i\), the convolutional output \(\mathbf{y}_i\) is computed as:
    \begin{equation}
    \mathbf{y}_i = \sum_{j} \mathbf{W}_j * \mathbf{x}_i + b_i
    \end{equation}
    where \(\mathbf{W}_j\) are the learnable convolutional kernels, \( * \) denotes the convolution operation, and \(b_i\) is a bias term. 
    
    This process generates feature maps capturing spatial hierarchies in the data, which are essential for modeling visual patterns and textures.

    \item \textbf{KAN-based Activation with Splines}: After convolution, each feature map \(\mathbf{y}_i\) is passed through an activation function defined by univariate spline transformations. This is formulated as:
    \begin{equation}
    \mathbf{z}_i = \sum_{k} w_k \phi_k(\mathbf{y}_i)
    \end{equation}
    where \(\phi_k(\mathbf{y}_i)\) are the spline-based activation functions applied to each convolved feature map, and \(w_k\) are the weights applied to each spline function.

    The spline functions in KAN are leveraged to achieve high non-linear representation, allowing CKAN to capture intricate dependencies and variations within local image features.
\end{enumerate}

CKAN integrates KAN’s nonlinear activation into convolutional processing, optimizing it for high-dimensional visual data with structured spatial dependencies. By coupling CNN’s spatial pattern extraction with KAN’s advanced spline-based nonlinearity, CKAN provides an expressive architecture for complex computer vision tasks.

\subsection{The Potential of KAN in Computer Vision}

\begin{figure*}[h!]
    \centering
    \includegraphics[width=\textwidth]{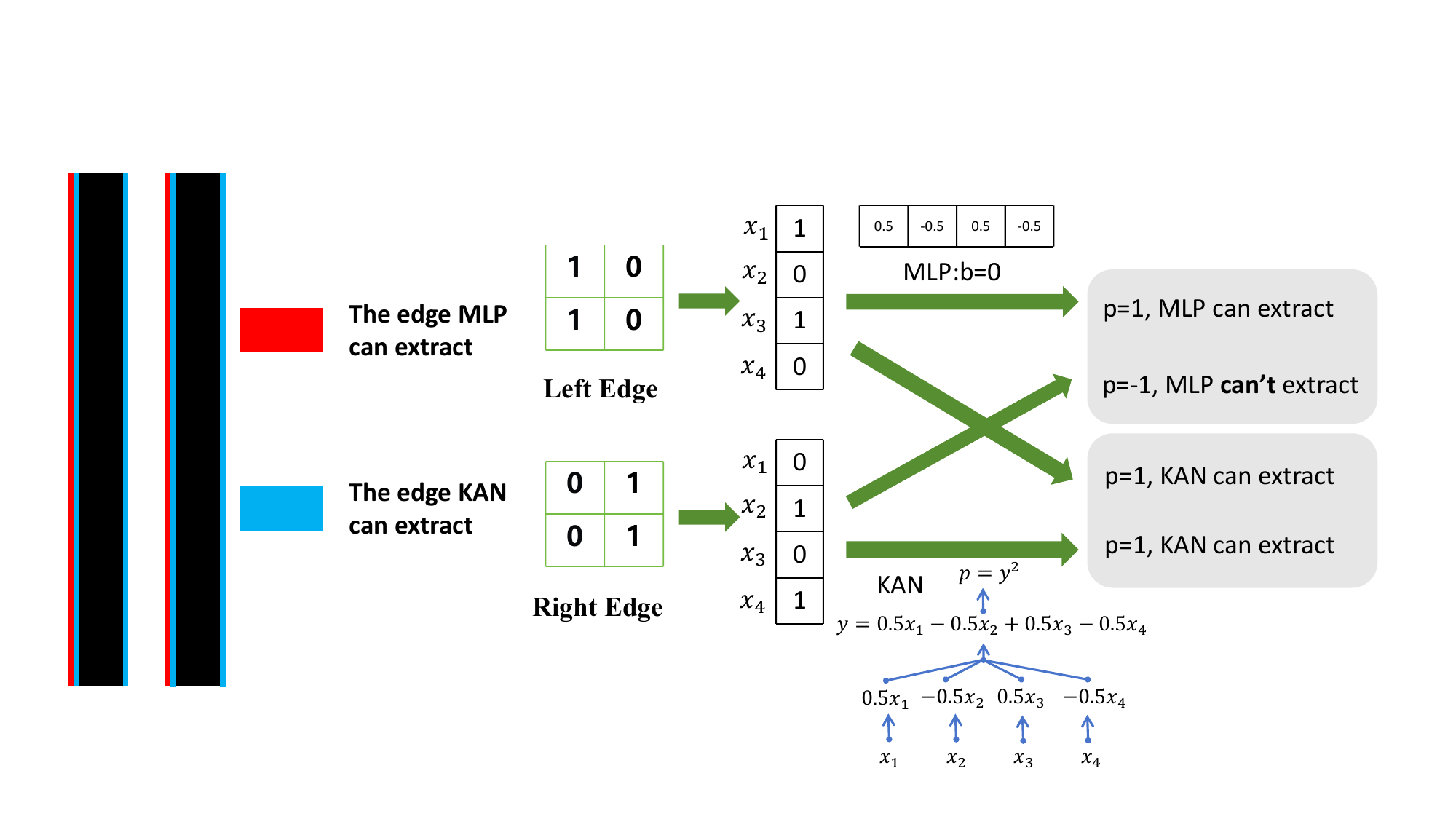}
    \captionsetup{singlelinecheck=false}
    \caption{\textbf{Illustration of KAN’s Potential in Edge Detection.} This figure presents an example comparing the edge detection capabilities of KAN and a traditional MLP. The MLP successfully detects the left edge pattern but struggles with the right edge, while KAN accurately identifies both patterns due to its nonlinear fitting capacity.}
    \label{fig:edge_detection_example}
\end{figure*}

As illustrated in Figure \ref{fig:edge_detection_example}, we present a simple edge detection example to demonstrate the potential of KAN in computer vision tasks. This example highlights the difference between KAN and a traditional MLP~\cite{rumelhart1986learning} in identifying specific edge patterns within an image.

In this scenario, the goal is to detect left and right edges within a set of pixels. For simplicity, we use a 4-pixel configuration, where the left edge is defined by a transition from a black (value of 1) to a white (value of 0) pixel, while the right edge shows the opposite transition, from white (value of 0) to black (value of 1).

In this example:
\begin{itemize}
    \item \textbf{Left Edge Detection with MLP:} An MLP with a single hidden layer can easily identify the left edge, as shown by the binary matrix in the upper half of the figure. Using a straightforward linear combination of the pixel values with specific weights and a bias term, the MLP correctly classifies the left edge.
    \item \textbf{Right Edge Detection with KAN:} When detecting the right edge pattern, however, an MLP struggles because of its linear structure, which lacks the necessary nonlinearity. By contrast, KAN, with its spline-based representation, can capture this pattern by adjusting segment parameters dynamically, enabling it to distinguish between subtle differences in pixel arrangements. As shown in the lower half of the figure, KAN accurately identifies the right edge due to its ability to fit more complex functions, making it better suited for this type of task.
\end{itemize}

This example illustrates that while MLPs are effective for simpler patterns, KAN’s nonlinear fitting ability allows it to capture more intricate visual features, such as edges of varying orientations or structures within an image. This capability is particularly valuable in computer vision, where data often contains complex spatial relationships, edges, and boundaries that require robust modeling beyond linear transformations.

By adapting to local patterns within an image, KAN demonstrates the potential to improve tasks like edge detection, object recognition, and segmentation. These initial results suggest that KAN, with proper regularization and techniques like Segment Deactivation to enhance generalization, could serve as a powerful tool in visual pattern recognition, leveraging its fitting strength to handle complex visual structures while minimizing sensitivity to noise.

\section{Experiments}

In this section, we present a series of experiments designed to evaluate the performance of Kolmogorov-Arnold Networks (KAN) and Convolutional KAN (CKAN) across several core computer vision tasks. These preliminary experiments help us assess the strengths and limitations of KAN in visual pattern recognition, providing a foundation for more in-depth analysis. Specifically, we apply KAN and CKAN to image classification and semantic segmentation tasks, using three widely-used datasets: CIFAR-10~\cite{krizhevsky2009learning}, CIFAR-100~\cite{krizhevsky2009learning}, and PASCAL VOC2012~\cite{everingham2010pascal}. Our goal is to provide an evaluation of KAN's capabilities in computer vision tasks through comprehensive experimentation. Subsequently, we will conduct an in-depth analysis and discussion based on the experimental results and consider potential ways to improve KAN's performance in computer vision tasks.

\subsection{Experiment 0: Preliminary Evaluation of KAN and CKAN on Core Vision Tasks}

\textbf{Image Classification.} We selected the CIFAR-100 dataset to conduct a preliminary experiment. MobileNetV2\cite{howard2017mobilenets,sandler2018mobilenetv2} was used as the baseline model, and we performed two replacement experiments: first, by substituting the final MLP layer with KAN for classification, and second, by replacing either the first or the final CNN layer in MobileNetV2 with CKAN. Due to the high memory requirements of CKAN, we limited our experiments to replacing only a single layer. To ensure a direct comparison between KAN and MLP, as well as CKAN and CNN, we used a minimal training configuration without any image augmentation or regularization techniques, employing cross-entropy as the loss function.

\textbf{Semantic Segmentation.} We selected the PASCAL VOC2012 dataset for our experiments, using UNet\cite{ronneberger2015u} as the baseline model. Since UNet does not include an MLP, we only replaced the final CNN layer used for classification with CKAN to conduct the experiment. Similarly, we adopted the simplest training configuration for this experiment. For all replacements in these experiments, we set the grid size to 5 and the spline order to 3.

\textbf{Result.} As shown in Table \ref{table1}, the experimental results indicate that models incorporating KAN and CKAN generally perform worse than the baseline models in classification and segmentation tasks, particularly the CKAN model. This result is unexpected, as the nonlinear characteristics of KAN and CKAN theoretically should enhance the model's fitting capability. This observation motivates further investigation into why the nonlinear fitting abilities of KAN and CKAN have not been fully utilized in computer vision tasks.

\begin{table*}[h!]
    \centering
    \captionsetup{justification=raggedright,singlelinecheck=false}
    \begin{tabular}{c|c|c|c}
        \hline
        \textbf{Task} & \textbf{Model} & \textbf{Replacement} & \textbf{Accuracy / mIoU (\%)} \\
        \hline
        \multirow{4}{*}{Image Classification} & MobileNet & Baseline & 61.14 \\
        & MobileNet & KAN (MLP replaced) & 60.98 \\
        & MobileNet & First Layer CKAN & 58.27 \\
        & MobileNet & Last Layer CKAN & 58.64 \\
        & MobileNet & two CKANs & 57.46 \\
        \hline
        \multirow{2}{*}{Semantic Segmentation} & UNet & Baseline & 63.28 \\
        & UNet & CKAN (Final CNN replaced) & 59.13 \\
        \hline
    \end{tabular}
    \caption{\textbf{Performance Comparison of Baseline Models with KAN and CKAN on CIFAR-100 and PASCAL VOC2012}.}
    \label{table1} 
\end{table*}

\subsection{Experiment 1: Effect of Dataset Size on KAN Performance}

\textbf{Setup.} In the following experiments, we designed three model architectures to evaluate the performance and potential of KAN and CKAN in visual tasks. First, we used a standard CNN model (CNN+MLP) as the baseline, which includes two convolutional layers followed by an MLP for classification. We then constructed two alternative models: one that replaces the first convolutional layer with CKAN (CNN+CKAN+MLP) and another that replaces the final MLP with KAN (CNN+KAN). The detailed configurations of these three models are shown in Table \ref{table2}.

\begin{table*}[h!]
    \centering
    \captionsetup{justification=raggedright,singlelinecheck=false}
    \begin{tabular}{l|l|l|l}
        \hline
        \textbf{Model} & \textbf{First Layer} & \textbf{Second Layer} & \textbf{Final Layer} \\
        \hline
        CNN+MLP & CNN (3$\rightarrow$32, 3x3) & CNN (32$\rightarrow$64, 3x3) & MLP \\
        CNN+CKAN+MLP & CKAN (3$\rightarrow$32, 3x3) & CNN (32$\rightarrow$64, 3x3) & MLP \\
        CNN+KAN & CNN (3$\rightarrow$32, 3x3) & CNN (32$\rightarrow$64, 3x3) & KAN \\
        \hline
    \end{tabular}
    \caption{\textbf{Experimental Model Configurations}. The baseline model (CNN+MLP) serves as a control for comparison with the CKAN and KAN-modified models.}
    \label{table2} 
\end{table*}

To verify whether KAN and CKAN indeed possess superior fitting capabilities, we designed the following experiments. Theoretically, KAN’s nonlinear spline functions should provide a fitting advantage over linear functions, although this was not clearly evident in the initial experiment. To further investigate, we conducted a series of experiments on CIFAR-10 using varying dataset sizes, ranging from 20\% to 100\% (20\%, 40\%, 60\%, 80\%, 100\%) of the available training data. In these experiments, we ensured balanced class distribution, with an equal number of samples from each class at every dataset size.

\textbf{Results.} The results, as shown in Table \ref{table3}, indicate that CNN + KAN consistently outperforms the baseline CNN + MLP across all dataset sizes, demonstrating KAN's stable fitting capability for this task. However, CKAN + CNN + MLP always performed worse than the baseline model, failing to exhibit the theoretical fitting advantage that CKAN should offer. Based on these experimental results, we can conclude that while KAN and CKAN theoretically possess stronger nonlinear fitting capabilities, in practice, CKAN has not effectively leveraged this advantage and performs worse than the traditional CNN + MLP. In contrast, KAN shows stable fitting performance, particularly as the dataset size increases, its performance improves. To investigate why CKAN failed to demonstrate its potential, we conducted Experiment 2.

\begin{table*}[h!]
    \centering
    \captionsetup{justification=raggedright,singlelinecheck=false}
    \begin{tabular}{c|c|c|c|c|c}
        \hline
        \textbf{Model} & \textbf{20\%} & \textbf{40\%} & \textbf{60\%} & \textbf{80\%} & \textbf{100\%} \\
        \hline
        CNN+MLP & 62.31\% & 67.26\% & 69.93\% & 71.33\% & 71.96\% \\
        CKAN+CNN+MLP & 61.82\% & 67.05\% & 69.03\% & 70.91\% & 71.88\% \\
        CNN+KAN & 64.02\% & 68.15\% & 70.41\% & 72.13\% & 72.58\% \\
        \hline
    \end{tabular}
    \caption{\textbf{Performance of Different Models on CIFAR-10 with Varying Dataset Sizes}. Each column shows the accuracy (\%) of each model as the dataset size increases from 20\% to 100\%.}
    \label{table3}
\end{table*}

\subsection{Experiment 2:} Based on the results of Experiment 1, we hypothesize that this may be due to the higher complexity of input in vision tasks, where image data is more likely to contain noise and outliers. The increased fitting capability may, therefore, lead to lower accuracy on the test set. To further investigate this issue, we designed Experiment 2, in which we intentionally introduce label noise to assess the performance of CKAN and KAN in noisy environments, with the aim of evaluating their robustness to noise in vision tasks.

\textbf{Setup.} In Experiment 2, we evaluated the robustness of different model architectures under varying levels of label noise. We gradually increased the noise from 10\% to 50\% and observed the impact on model performance. The three models tested were CNN as the baseline, CKAN+CNN+MLP with CKAN replacing the first convolutional layer, and CNN+KAN with KAN replacing the final MLP layer.

\textbf{Results.} As shown in Table \ref{table4}, as the noise level increases, the accuracy of all models decreases, with CKAN+CNN+MLP and CNN+KAN showing a more pronounced decline compared to the baseline CNN model. Although the initial performance of the KAN and CKAN models is slightly better than the baseline under low noise conditions, they exhibit significantly higher sensitivity to noise at higher noise levels. This indicates that KAN and CKAN have poor robustness, and their strong fitting capabilities do not seem suitable for handling high-complexity image inputs in computer vision tasks. Instead, they may lead to overfitting, thereby reducing the models' performance in real-world applications.

\begin{table*}[h!]
    \centering
    \captionsetup{justification=raggedright, singlelinecheck=false}
    \begin{tabular}{c|c|c|c|c|c}
        \hline
        \textbf{Model} & \textbf{10\%} & \textbf{20\%} & \textbf{30\%} & \textbf{40\%} & \textbf{50\%} \\
        \hline
        CNN+MLP & 69.13\% & 67.93\% & 65.38\% & 63.88\% & 62.87\% \\
        CKAN+CNN+MLP & 69.59\% & 66.73\% & 64.19\% & 62.56\% & 58.43\% \\
        CNN+KAN & 69.95\% & 67.81\% & 64.33\% & 62.84\% & 59.20\% \\
        \hline
    \end{tabular}
    \caption{\textbf{Performance of Different Models with Increasing Label Noise}. Each column represents the accuracy (\%) as the label noise increases from 10\% to 50\%.}
    \label{table4}
\end{table*}

\subsection{Experiment 3: Impact of Regularization on the Generalization of KAN and CKAN}

In Experiment 2, we observed that noise significantly impacts the performance of KAN and CKAN models. Based on this finding, Experiment 3 aims to study the effect of regularization on the generalization ability of KAN and CKAN. We introduced L1 regularization~\cite{tibshirani1996regression} in the models and tested various regularization parameters to observe its influence on performance in visual tasks. This experiment evaluates whether regularization can effectively enhance the stability and robustness of KAN and CKAN models.

\textbf{Setup.} To investigate the impact of L1 regularization on the generalization ability of KAN and CKAN, we conducted two experiments with different data conditions:
\begin{itemize}
    \item \textbf{Noisy Environment.} We applied four L1 regularization strengths (0, 0.0001, 0.001, 0.01) on a dataset with 30\% label noise.
    \item \textbf{Normal Environment with Limited Data.} We repeated the experiment on a clean 60\% subset of the dataset, testing the same regularization strengths.
\end{itemize}

\textbf{Results.} As shown in Tables \ref{table5} and \ref{table6}, we observed that with moderate L1 regularization, models incorporating KAN outperformed the baseline CNN model in accuracy, while models with CKAN consistently performed worse. This indicates that KAN's fitting capability was better utilized, with reduced sensitivity to noise and outliers, highlighting the significant role of regularization in enhancing its generalization ability. In contrast, even with regularization, CKAN did not surpass the baseline model, and due to its high memory usage and slow computation speed, we concluded that the current CKAN structure has limited practical potential and is not a suitable alternative to CNNs. Based on these findings, we decided to abandon further exploration of CKAN and focus on improving KAN in the subsequent experiments, aiming to optimize its performance as the final layer of the model while enhancing its robustness.

\begin{table*}[h!]
    \centering
    \captionsetup{justification=raggedright, singlelinecheck=false}
    \begin{tabular}{c|c|c|c|c}
        \hline
        \textbf{Model} & \textbf{0} & \textbf{0.0001} & \textbf{0.001} & \textbf{0.01} \\
        \hline
        CNN+MLP & 65.38\% & 65.54\% & 65.79\% & 63.47\% \\
        CKAN+CNN+MLP & 64.19\% & 64.98\% & 65.64\% & 63.02\% \\
        CNN+KAN & 64.33\% & 65.26\% & 66.03\% & 63.94\% \\
        \hline
    \end{tabular}
    \caption{\textbf{Effect of L1 Regularization on Model Accuracy with 30\% Label Noise}. Each column shows the accuracy (\%) of each model with increasing L1 regularization values.}
    \label{table5}
\end{table*}

\begin{table*}[h!]
    \centering
    \captionsetup{justification=raggedright, singlelinecheck=false}
    \begin{tabular}{c|c|c|c|c}
        \hline
        \textbf{Model} & \textbf{0} & \textbf{0.0001} & \textbf{0.001} & \textbf{0.01} \\
        \hline
        CNN+MLP & 69.93\% & 70.01\% & 70.23\% & 67.58\% \\
        CKAN+CNN+MLP & 69.03\% & 69.54\% & 70.12\% & 67.32\% \\
        CNN+KAN & 70.41\% & 70.98\% & 71.56\% & 63.94\% \\
        \hline
    \end{tabular}
    \caption{\textbf{Effect of L1 Regularization on Model Accuracy with 60\% Dataset Size}. Each column shows the accuracy (\%) of each model with increasing L1 regularization values.}
    \label{table6}
\end{table*}

\subsection{Experiment 4: Enhanced Regularization and Segment Deactivation for Improved KAN Generalization}

Based on the results of Experiment 3, we found that improving the generalization ability of KAN and reducing its sensitivity to noise is crucial. To address this, we propose Smoothness Regularization method and introduce the Segment Deactivation technique, aiming to enhance the robustness and generalization of KAN.

\textbf{Smoothness Regularization Method.} To enhance the generalization ability of KAN and reduce its sensitivity to noise, we introduce a smoothness regularization method tailored to KAN's spline-based structure. Specifically, this regularization penalizes abrupt changes in the spline's behavior, encouraging smooth transitions between segments and preventing overfitting to noisy data. The smoothness regularization term is formulated as follows:

\begin{equation}
    \mathcal{R}_{\text{smooth}} = \lambda \sum_{i=1}^{N} \int \left( \frac{d^2 S_i(x)}{dx^2} \right)^2 dx,
\end{equation}

where \( S_i(x) \) is the spline function for segment \( i \), \( N \) is the total number of spline segments, and \( \lambda \) is the regularization strength. This term penalizes the second derivative of each spline segment, promoting a smoother overall function.

This smoothness regularization leverages KAN’s spline-based structure, improving its stability across segments while maintaining strong fitting capability and mitigating the impact of noise.

\textbf{Segment Deactivation Technique.} The Figure To further improve the robustness and generalization of KAN, we propose a novel regularization technique termed \textit{Segment Deactivation}. Similar to Dropout~\cite{srivastava2014dropout}, this approach selectively deactivates specific spline functions, reducing them to linear functions with a certain probability only during the training phase. By simplifying the entire spline function into a straight line in selected cases, Segment Deactivation helps prevent overfitting and mitigates sensitivity to noise. During testing, all spline functions operate in their original nonlinear form, allowing the model to perform with its full fitting capacity.

\begin{figure}[h!]
    \centering
    \includegraphics[width=0.5\textwidth]{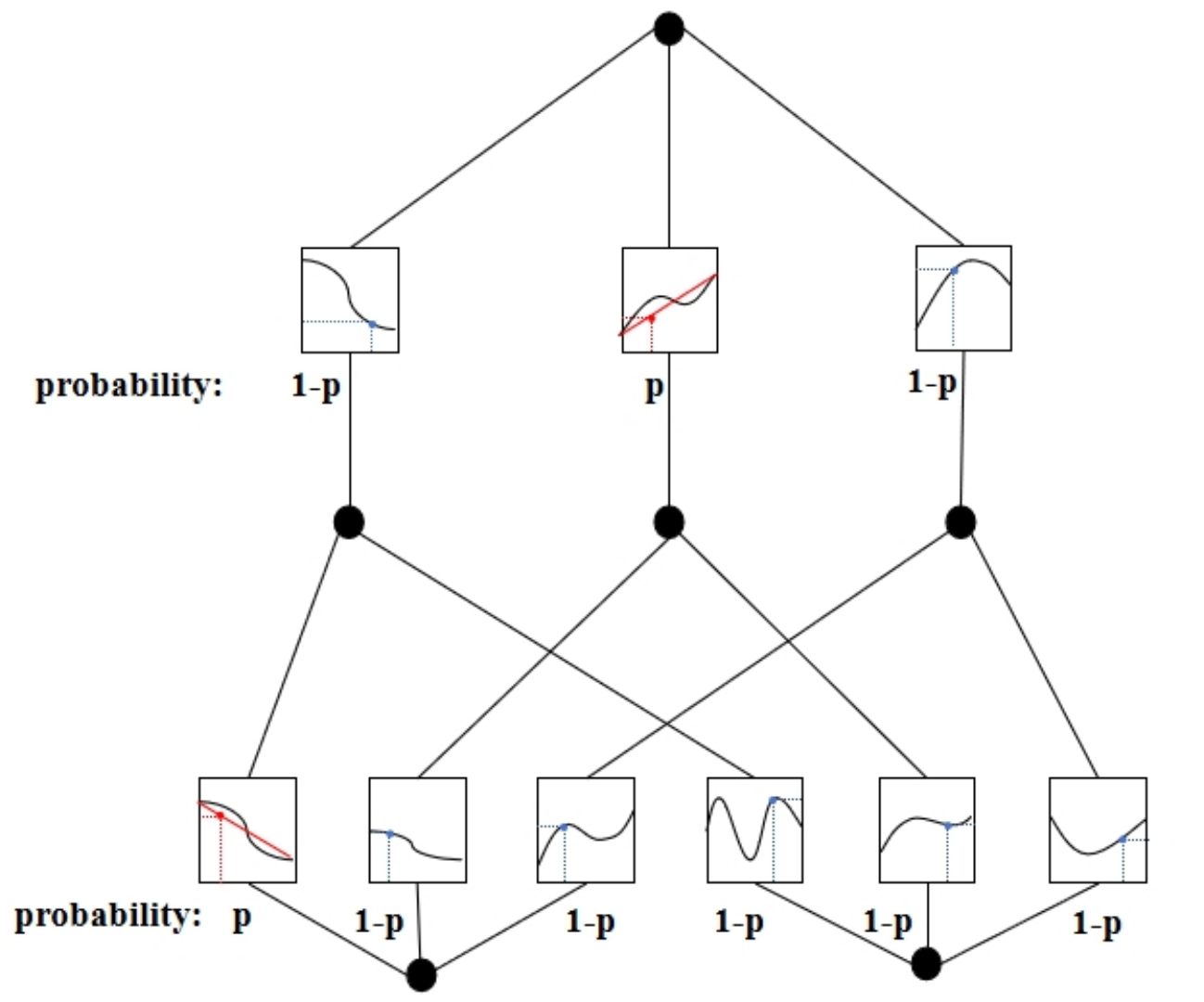}
    \captionsetup{singlelinecheck=false}
    \caption{\textbf{Illustration of Segment Deactivation Technique}. This figure demonstrates the simplification process where the entire spline function between its start and end points is replaced with a linear connection (shown in red) with a certain probability \( p \). This technique reduces complexity in noisy segments, enhancing generalization.}
    \label{fig:segment_deactivation_example}
\end{figure}

The Segment Deactivation regularization is implemented as follows:
\begin{equation}
    S(x) = 
    \begin{cases} 
      S(x) & \text{with probability } 1 - p, \\
      a \cdot x + b & \text{with probability } p,
    \end{cases}
\end{equation}
where \( S(x) \) is the original spline function for a specific input, and \( a \) and \( b \) are coefficients that define a line connecting the start and end points of the entire spline function. The probability \( p \) controls the likelihood of deactivation, balancing between preserving the spline's expressiveness and encouraging generalization.The simplifie Segment Deactivation Technique process is illustrated in Figure \ref{fig:segment_deactivation_example}.

By introducing controlled linearity over the entire spline function, Segment Deactivation reduces the risk of overfitting to noise and enhances KAN’s generalization capability. This technique is particularly beneficial in noisy environments, as it forces the model to learn more generalized patterns across the data.

\begin{table*}[h!]
    \centering
    \captionsetup{justification=raggedright, singlelinecheck=false}
    \begin{tabular}{l|c}
        \hline
        \textbf{Model Configuration} & \textbf{Accuracy (\%)} \\
        \hline
        CNN + MLP & 71.96 \\
        CNN + KAN & 72.58 \\
        CNN + KAN + Smoothness Regularization Method & 73.04 \\
        CNN + KAN + Segment Deactivation Technique & 73.12 \\
        CNN + KAN + Smoothness Regularization Method + Segment Deactivation Technique & 73.30 \\
        \hline
    \end{tabular}
    \caption{\textbf{Performance Comparison of CNN with KAN, Smoothness Regularization, and Segment Deactivation Techniques on CIFAR-10 Dataset}. Each row shows the accuracy (\%) for different configurations of CNN with the proposed methods on CIFAR-10.}
    \label{table7}
\end{table*}

\begin{table*}[h!]
    \centering
    \captionsetup{justification=raggedright, singlelinecheck=false}
    \begin{tabular}{l|c}
        \hline
        \textbf{Model Configuration} & \textbf{Accuracy (\%)} \\
        \hline
        MobileNet + NLP & 61.14 \\
        MobileNet + KAN & 60.98 \\
        MobileNet + KAN + Smoothness Regularization Method & 62.03 \\
        MobileNet + KAN + Segment Deactivation Technique & 62.37 \\
        MobileNet + KAN + Smoothness Regularization Method + Segment Deactivation Technique & 63.12 \\
        \hline
    \end{tabular}
    \caption{\textbf{Performance Comparison of MobileNet with KAN, Smoothness Regularization, and Segment Deactivation Techniques on CIFAR-100 Dataset}. Each row shows the accuracy (\%) for different configurations of MobileNet with the proposed methods on CIFAR-100.}
    \label{table8}
\end{table*}

We conducted experiments on both CIFAR-10 and CIFAR-100 datasets to evaluate the effectiveness of the proposed Smoothness Regularization method and Segment Deactivation technique in improving the generalization of KAN. For each dataset, we tested four configurations: KAN alone, KAN with Smoothness Regularization, KAN with Segment Deactivation, and KAN with both techniques combined. This ablation study allows us to examine the individual and combined contributions of each method to the model's accuracy.

\textbf{Results.} As shown in Tables \ref{table7} and \ref{table8}, both the Smoothness Regularization method and Segment Deactivation technique resulted in notable accuracy improvements when added to the KAN-enhanced models on both CIFAR-10 and CIFAR-100. In particular, the configuration combining both methods achieved the highest accuracy, demonstrating that the proposed techniques effectively mitigate overfitting and improve generalization. Furthermore, these results indicate that each method contributes positively to KAN’s robustness, and when used together, they complement each other to further enhance performance.

\section{Conclusions}

This study explores the application of Kolmogorov-Arnold Networks (KAN) and their variants in computer vision tasks. Despite KAN's strong fitting capabilities in theory, our experimental results indicate that the high complexity of visual data and the prevalence of noise pose significant challenges to its generalization ability. This is particularly evident in the case of Convolutional KAN, which performs worse than traditional CNNs and is not suitable as a replacement for conventional CNNs. Similarly, directly replacing the NLP layer with KAN did not result in significant performance improvements across most tasks. To address these issues, we propose a smoothness regularization method and the "Segment Deactivation" technique. These methods effectively reduce overfitting to noise and promote model smoothness, significantly enhancing the robustness of KAN. Ultimately, KAN achieves better performance when replacing the final NLP layer of the model, showing some potential for future applications.

{
    \small
    \bibliographystyle{ieeenat_fullname}
    \bibliography{main}
}


\end{document}